\documentclass[lettersize,journal]{IEEEtran}
\usepackage{amsmath,amsfonts}
\usepackage{algorithmic}
\usepackage{algorithm}
\usepackage{array}
\usepackage[caption=false,font=normalsize,labelfont=sf,textfont=sf]{subfig}
\usepackage{textcomp}
\usepackage{stfloats}
\usepackage{url}
\usepackage{verbatim}
\usepackage{graphicx}
\usepackage{cite}
\usepackage{booktabs}
\usepackage{tabularx}
\usepackage{multirow}
\usepackage{hyperref}
\usepackage[shortlabels]{enumitem}
\begin{document}
\title{Data-Driven Multi-step Nonlinear Model Predictive \\ Control for Industrial Heavy Load Hydraulic Robot}

\author{Dexian Ma, and Bo Zhou, \IEEEmembership{Member, IEEE}
\thanks{This work was supported in part by the National Natural Science Foundation (NNSF) of China under Grant 62073075. \textit{(Corresponding author: Bo Zhou.)}}
\thanks{D. X. Ma and B. Zhou are with the Key Laboratory of Measurement and Control of Complex Systems of Engineering (School of Automation, Southeast University), Ministry of Education, Nanjing 210096, China (e-mail: {mdx@seu.edu.cn}, {zhoubo@seu.edu.cn},). }}

\markboth{}%
{Shell \MakeLowercase{\textit{et al.}}}

\IEEEpubid{}

\maketitle

\begin{abstract}
Automating complex industrial robots requires precise nonlinear control and efficient energy management. This paper introduces a data-driven nonlinear model predictive control (NMPC) framework to optimize control under multiple objectives. To enhance the prediction accuracy of the dynamic model, we design a single-shot multi-step prediction (SSMP) model based on long short-term memory (LSTM) and multilayer perceptrons (MLP), which can directly obtain the predictive horizon without iterative repetition and reduce computational pressure. Moreover, we combine offline and online models to address disturbances stemming from environmental interactions, similar to the superposition of the robot's free and forced responses. The online model learns the system's variations from the prediction mismatches of the offline model and updates its weights in real time. The proposed hybrid predictive model simplifies the relationship between inputs and outputs into matrix multiplication, which can quickly obtain the derivative. Therefore, the solution for the control signal sequence employs a gradient descent method with an adaptive learning rate, allowing the NMPC cost function to be formulated as a convex function incorporating critical states. The learning rate is dynamically adjusted based on state errors to counteract the inherent prediction inaccuracies of neural networks. The controller outputs the average value of the control signal sequence instead of the first value. Simulations and experiments on a 22-ton hydraulic excavator have validated the effectiveness of our method, showing that the proposed NMPC approach can be widely applied to industrial systems, including nonlinear control and energy management. 
\end{abstract}


\section{Introduction}\label{Introduction}
\IEEEPARstart{T}{he} shift towards automation and unmanned operations in industrial systems is a crucial strategy for enhancing productivity and can lead to significant economic benefits \cite{zhang2021autonomous}. Most actuating components in these systems are manipulators or other mechatronic-hydraulic products. Achieving high-precision control over these robots is vital for successfully executing engineering tasks. Several traditional methods are commonly used for process and robotic control, including Proportional-Integral-Derivative (PID) control \cite{malki1997fuzzy},\cite{premkumar2015fuzzy},\cite{van2020self} and model-based control \cite{kim2019modeling},\cite{yao2017active},\cite{williams2017information},\cite{nguyen2020global}. PID control is widely adopted due to its straightforward principles and broad applicability, meeting the essential requirements of many industrial applications. However, it has limitations when applied to systems with complex nonlinearities or when optimization is necessary, such as in high-order time-delay systems or resource allocation tasks. Model-based control methods, particularly Model Predictive Control (MPC), are well-known for achieving optimal control in nonlinear systems \cite{dai2020robust},\cite{di2013stochastic}. MPC uses a dynamic model to predict system behavior over a future time horizon at each sampling interval and calculates a sequence of optimal control signals to enhance system performance \cite{shen2017trajectory}. The effectiveness of MPC relies on the accuracy of the system's dynamics model, which presents considerable challenges. Nonlinear factors such as combustion \cite{van2001modeling}, friction \cite{xiao2021sensorless}, and fluid dynamics \cite{gholizadeh2014modeling} are difficult to measure directly and often lack precise physical representations. Additionally, the dynamics model of a robot may change when it interacts with its external environment. Typical scenarios include handling different loads or the increasing temperature of the robot's drive motors or hydraulic oil \cite{gao2018oscillatory}. Moreover, the receding horizon approach in MPC and complex dynamics may lead to myopic predictions \cite{mitsioni2023safe}, potentially resulting in control failures.

Data-driven approaches provide a new perspective for modeling and controlling robotic systems. This method primarily involves using neural networks or Gaussian process regression (GPR) to learn both forward and inverse dynamics, thus avoiding the inaccuracies commonly associated with direct physical modeling. Inverse dynamics \cite{kim2021force} can be used for direct control or feedforward compensation, while forward dynamics is crucial in model reference adaptive control (MRAC) or MPC \cite{cui2022approximate},\cite{wang2020parameter}. The computational complexity of GPR is influenced by the training set due to the non-parametric nature, necessitating additional preprocessing to select an appropriate subset of data  \cite{ambikasaran2015fast}. In contrast, neural networks are generally more intuitive and user-friendly, which has led to their widespread adoption in the field. A data-driven nonlinear model predictive control (DD-NMPC) approach combines neural network learning models with MPC. By leveraging the universal approximation properties of neural networks, it is possible to learn the dynamics of the control plant either offline or online, resulting in a highly predictive nonlinear model that can be used to optimize the control sequence. However, DD-NMPC also encounters several challenges as the follows.

\textit{Accurate predictive modeling}: The effectiveness of predictive models relies not only on the extensive collection of data but is also heavily influenced by the design of the neural network. For example, recurrent neural networks (RNNs) are more effective than multilayer perceptrons (MLPs) for predicting time-series data because they can capture temporal dependencies \cite{huang2022lstm}. Additionally, the predictive horizon in NMPC needs to be updated continuously through iterations, which can be affected by fitting errors in the neural networks. As prediction errors accumulate and amplify over multiple iterations, they can ultimately reduce the control system's effectiveness.
	
\textit{Environmental Interaction}: The dynamics of a robot can change when it interacts with the external environment, which may reduce the prediction accuracy of a learning model that has been trained on offline data. Online or active learning methods aim to improve offline learning by continuously updating the model with new data collected during operation \cite{saviolo2023active}. However, this iterative process can be time-consuming and risky, and the potential for non-convergence can make it impractical to implement in engineering tasks.
	
\textit{Real-time Optimization}: In linear MPC, determining control actions is addressed as a quadratic programming (QP) or sequential quadratic programming (SQP) problem. This involves using a quadratic cost function, which efficiently identifies either the optimal or a locally optimal solution. On the other hand, when working with nonlinear models, each iteration requires solving a nonlinear programming (NLP) problem. While there is flexibility in the form of the cost function, which can be adapted to various convex functions to meet specific needs, the direct application of linear MPC methods is often impractical.

\subsection{Contributions}\label{Introduction-A}
Inspired by online learning and nonlinear programming, we develop a novel learning-based single-shot multi-step prediction (SSMP) model paired with a corresponding NMPC strategy to tackle the three challenges outlined earlier in this work. Unlike traditional methods that incrementally construct a predictive horizon based on single-step predictions, our approach promotes direct multi-step prediction. This method effectively reduces the accumulation of errors and results in a more accurate model. Additionally, the SSMP framework minimizes the computational demands typically associated with neural networks. Our proposed predictive model leverages the strengths of both offline and online learning, effectively alleviating the burdens of online learning by building upon the robust foundation established during offline learning. This strategy enhances the model’s adaptability to dynamic environmental changes and significantly lowers the risk of uncontrolled behavior, which is a common issue in systems relying solely on online learning. We also account for approximation errors related to learning models within the NMPC. An adaptive parameter tuning method has been designed to address disturbances arising from model uncertainties, thus allowing us to obtain an optimized control sequence. Compared to state-of-the-art solutions, our approach is distinguished by its practical and economic advantages, particularly in applications such as the unmanned control of heavy-load hydraulic excavators. In summary, the main contributions of this work are as follows:
\begin{enumerate}[nosep]
	\item Designing an SSMP neural network model based on long short-term memory (LSTM) and MLP to learn the nonlinear dynamics of robots. By predicting changes in states rather than the direct states, a marked enhancement can be achieved in the precision of the predictive model. The method for collecting data is also discussed.
	
	\item Combining offline and online models when robots interact with the external environment. Rather than directly modifying the parameters of the offline model, we employ a new MLP to learn and compensate for the mismatches. This approach helps us avoid the risk of losing control during interactions.
	
	\item Proposing an adaptive gradient descent optimization method for NMPC. We detail a cost function designed for multi-objective optimization and elucidate the derivation process for the predictive model. The impact of model fitting errors can be reduced by introducing the variable learning rate. Moreover, we use the average value of the control sequence instead of the first element.
	
	\item Validating the proposed methodology through simulation and experimentation on a heavy-load hydraulic excavator. The results demonstrate that the predictive model can adapt to complex nonlinear systems under various load conditions and achieve economic efficiency while meeting work requirements.
\end{enumerate}

\section{Related Works}\label{Related Works}
The proposed method involves learning-based dynamics modeling and optimization for NMPC. We will begin with a brief overview of current learning models and training methods, followed by a discussion on solution methods for learning-based NMPC.

\textit{Learning-based nonlinear dynamics model}: Neural networks have emerged as a cutting-edge approach for capturing robotic systems' forward and inverse dynamics due to their universal approximation property. Our focus is on the neural network structures typically used for modeling forward dynamics, which include MLP \cite{cheng2015neural}, deep belief networks (DBN) \cite{wang2020deep}, radial basis function networks (RBF) \cite{chen2019rbfnn}, physics-informed neural networks (PINN) \cite{karniadakis2021physics}, LSTM networks \cite{huang2022lstm}, and echo state networks (ESN) \cite{jordanou2021echo}, among others. These learning models usually take the system's state variables and input signals as inputs, with data acquired through sensors. Most of these models are designed to learn dynamics in discrete time intervals. Provided that a sufficient and high-quality dataset is used, this approach keeps the model's predictive capability intact \cite{saviolo2023active}. The output of these models is flexible and can be tailored to meet specific needs. For example, a complete nonlinear system can be treated as a black box, allowing the neural network to fit the entire dynamics. Alternatively, prior knowledge about the physical mechanisms can be integrated to model part of the dynamics, which is commonly done in PINN. Unlike feedforward structures like MLP, RNNs are more complex and excel at extracting features from historical data, making them advantageous for time-series prediction and higher-order or time-delay nonlinear systems. Many predictive models provide single-step predictions, which can be extended iteratively to form a predictive horizon. \cite{huang2022lstm} suggests using LSTM for multi-step predictions; however, it does not specify the input and output details. In response, we have combined the strengths of LSTM and MLP to create the SSMP model, accompanied by a detailed discussion of its structure. The LSTM-MLP leverages the enhanced predictive capabilities of LSTM as an offline model, while MLP is employed as the online model due to its simpler training process.

\textit{Learning method}: Supervised training is typically used for the regression process, with the cost function defined as the norm of the prediction error \cite{lee2022precision}. Offline learning utilizes existing demonstration datasets, allowing for training without physical risks and minimizing the need for computational resources. However, this approach relies on high-quality datasets. When robots interact with their external environment, it is crucial to thoroughly consider the state across nearly all dimensions to capture the dynamic progression. Online learning \cite{bechtle2021leveraging},\cite{waegeman2012feedback} enables the synchronous updating of neural networks with refreshed data, but the learning process is fraught with uncertainties that can lead to control instability in real-time applications, particularly in robots that operate with high power and significant loads. \cite{saviolo2023active} has integrated offline learning from past experiences with online learning to address these challenges as robots engage with unknown environments. This approach allows for model refinement by adjusting the weights of the last layer of the neural network through active learning. Our approach differs from traditional practices by conceptualizing the model as a blend of offline and online elements, similar to superimposing a robot's free and forced responses. This framework allows the system to quickly adapt to interactions, such as when a robot picks up an object. We effectively minimize associated risks by using the non-interactive state as a reference for action selection within NMPC.

\textit{NMPC and optimization}: NMPC primarily employs nonlinear predictive models and varies in solution methods. A common approach is to utilize linear MPC, which approximates the nonlinear objective function and constraints at each iteration as linear or quadratic functions. These approximations are then solved using QP or SQP. \cite{yan2012model} employs RNNs as predictive models for the control plant, linearizing them into state-space systems for MPC to calculate the control signal. \cite{jordanou2021echo} separates nonlinear dynamic models into free and forced responses through first-order Taylor expansion. The ESN is utilized for free response and the forced response is obtained by a fast and recursive calculation of the input-output sensitivities from the ESN. While this approach is intuitive, the linearization process and repeated iterations can lead to a loss of predictive accuracy and increased demand for computational resources.
An alternative approach to calculating the control sequence involves directly optimizing the cost function, utilizing evolutionary computation and gradient-based methods. Evolutionary computation methods, such as stochastic shooting \cite{saviolo2023active} and differential evolution optimization \cite{zhang2021nonlinear}, effectively manage complex constraints. However, these methods can be impractical for real-world applications due to their high computational resource requirements and time constraints. In contrast, direct computation of gradients within neural networks offers a more feasible solution, with optimization typically achieved through GD or the levenberg-marquardt (LM) algorithm, both of which are more universally applicable. When using gradient-based methods, setting a termination precision or a maximum number of iterations is expected to ensure a suboptimal solution. In \cite{cheng2015neural}, an MLP is used to realize a nonlinear autoregressive moving average with exogenous inputs (NARMAX) model for a piezoelectric actuator. The control sequence is obtained by solving the cost function using the LM algorithm. \cite{wang2020deep} designs a growing deep belief network that directly utilizes the control signals u(t) as the input and optimizes through GD. 
The final approach is the learning-based explicit NMPC \cite{drgovna2022differentiable},\cite{kang2022tracking}, where the optimal control sequence is computed through a neural network. The optimization process does not require iterations, significantly reducing the demand for computational resources. However, the offline-trained model lacks the flexibility to adapt to different tasks and accommodate variations in environmental interactions. 
In summary, directly computing the derivatives of neural networks and employing GD-based optimization methods require less computation and can achieve satisfactory results.

\section{Preliminaries and Application Objects}\label{Preliminaries and Application Objects}
\subsection{Preliminaries}\label{Preliminaries and Application Objects-A}
Table \ref{datasettable} summarizes the symbols used in this paper along with their respective meanings. Some symbols may have multiple meanings, but their application within the model remains consistent. For a continuous system, we have utilized a zero-order hold on the input u to represent the discretization of the system dynamics.

\begin{table}[t]
\begin{center}
	\caption{Notation Used in This Work}\label{datasettable}
	\begin{tabular}{cl}
		\toprule
		Symbol 				& Meaning \\
		\midrule
		$\mathbb{R}$ 		&  The set of real numbers\\
		$X$          		&  State vector\\
		$U$          		&  Input vector\\
		$Y$          		&  Output vector\\
		$f$          		&  Discrete-time dynamics\\
		$S$          		&  Constraints on the dynamics model\\
		${{R}_{x}}$     	&  Reference of $x$\\
		$q$ 				&  Joint angle\\
		$\omega$        	&  Engine’s rotational speed\\
		$Q$          		&  Flow rate\\
		$L$          		&  Displacement\\
		${}^{i}{{w}_{x}}$   &  The $i$ layer’s weight constants of neural networks $x$\\
		${}^{i}{{B}_{x}}$   &  The $i$ layer’s bias constants of neural networks $x$\\
		$n$     			&  State size\\
		$m$          		&  Input size\\
		$l$          		&  Output size\\
		$h$          		&  Number of historical data\\
		$N$          		&  NMPC predictive horizon\\
		$\eta$          	&  Learning rate\\
		\bottomrule
	\end{tabular}
\end{center}
\end{table}

\subsection{Application Objects}\label{Preliminaries and Application Objects-B}
\begin{figure}[!t]
	\centering
	\includegraphics[width=3.2in]{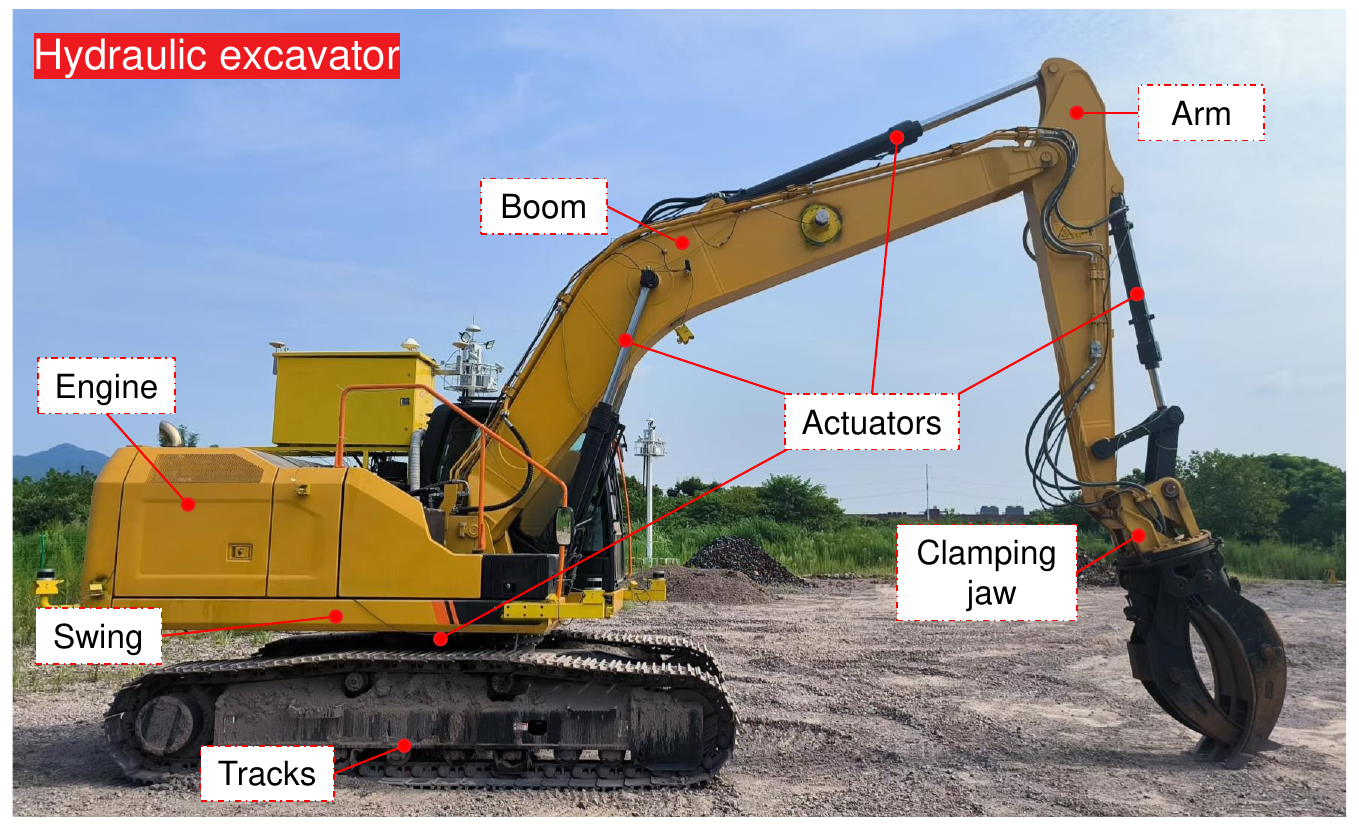}
	\caption{Power source, joint and actuation system of hydraulic excavator. The final joint requires selecting the appropriate tool hand based on the task. To adapt to grasping tasks, we change to a clamping jaw.}
	\label{hydraulic_excavator_introduction}
\end{figure}

Our application object is a hydraulic excavator, a heavy-duty mobile manipulation system powered by diesel fuel. This equipment is commonly used in construction, mining, waste management, and other heavy-load operations. With its excellent power-to-weight ratio and proven reliability, the hydraulic excavator currently has no viable substitutes, ensuring its continued significance in the industry for the foreseeable future. The hydraulic excavator is a complex assembly that comprises the following systems, as shown in Fig.\ref{hydraulic_excavator_introduction}.

\textit{Power System}: The power system of a hydraulic excavator is driven by a diesel engine. This engine operates piston pumps that deliver hydraulic fluid. It features near-constant torque regulation, which means the engine's power output is reflected in variations in speed, while the torque output remains relatively stable. As the throttle opening increases, the engine's output power also rises correspondingly.

\textit{Hydraulic Drive System}: The high-pressure oil produced by the hydraulic pump is transported to the actuator system through pipelines. To reduce fuel consumption and ensure optimal power matching, negative or positive flow hydraulic circuits are commonly used.

\textit{Valve System}: The direction or pressure of hydraulic fluid is adjusted through proportional solenoid valves, relief valves, and others.

\textit{Actuation System}: Including the left and right travel motors, the swing rotation motor, boom hydraulic cylinders, arm hydraulic cylinders, and bucket hydraulic cylinders, among others.

\textit{Electronic Component System}: It is responsible for converting signals into actions for the underlying electronic components. Furthermore, manufacturers typically incorporate basic functions in commercial products, such as low-power mode.

The unmanned operation of hydraulic excavators significantly reduces costs and eliminates the risk of injury to operators, thereby enhancing efficiency and safety. For a robot to perform engineering tasks automatically, it is essential to effectively control the power and actuation system. However, achieving high-quality automatic control of hydraulic excavators is a complex challenge. This complexity arises from the inherent nonlinearity of diesel engine and hydraulic drive systems and the strong coupling between the oil pump and the actuator within the positive flow system. This scenario involves nonlinear control and challenges related to power matching and energy scheduling. In addition, commercial hydraulic excavator products involve trade secrets, making it difficult to obtain complete structural information, which prevents the establishment of an accurate mathematical model. Therefore, we aim to address the unmanned control issue using data-driven methods, a currently prevalent approach in the field.

\section{Methodology}\label{Methodology}
In the following, we present the SSMP offline neural network model constructed utilizing LSTM and MLP, which is informed by Taylor expansion. We propose a prediction framework combining the offline and online models for enhanced performance. Finally, we discuss the cost function and optimization methods of NMPC.

\subsection{Offline Learning Model for Nonlinear Dynamics}\label{Methodology-A}
Consider a discrete-time multi-input and multi-output (MIMO) nonlinear system with time delays. Define the state vector $X\in {{\mathbb{R}}^{n}}$, control input $U\in {{\mathbb{R}}^{m}}$, and output vector $Y\in {{\mathbb{R}}^{l}}$. At every time step $t$, the system can be described as:
\begin{equation}\label{eq1}
\left\{ \begin{aligned}
	& {{X}_{t+1}}=f(X_{t}^{h},U_{t}^{h}) \\ 
	& {{Y}_{t+1}}=C\cdot {{X}_{t+1}} \\ 
\end{aligned} \right.
\end{equation}
where $X_{t}^{h}=\left[ \begin{matrix}
	{{X}_{t-h}} & {{X}_{t-h+1}} & \ldots  & {{X}_{t}}  \\
\end{matrix} \right]$, 
$U_{t}^{h}=\left[ \begin{matrix}
{{U}_{t-h}} & {{U}_{t-h+1}} & \ldots  & {{U}_{t}}  \\
\end{matrix} \right]$. 
$C$ is a constant matrix and $f$ is the system dynamics that must be identified. Future states are generally predicted using rolling iteration that relies on the current state and inputs. However‌, our primary focus is developing appropriate neural network learning models that can directly predict future states over a specified time frame, utilizing only the current state and inputs. Avoiding rolling predictions reduces error accumulation and lessens the computational resources needed for NMPC.

Define $\Delta X_{t+i}^{h}=X_{t+i}^{h}-X_{t-1}^{h}$, $\Delta U_{t+i}^{h}=U_{t+i}^{h}-U_{t-1}^{h}$, then we can obtain:
\begin{equation}\label{eq2}
	\Delta X_{t+1}^{h}={{f}^{h}}(X_{t}^{h},U_{t}^{h})-X_{t-1}^{h}
\end{equation}
by conducting a Taylor expansion of (\ref{eq2}) at $(X_{t-1}^{h},U_{t-1}^{h})$, it can be rewritten as:
\begin{equation}\label{eq3}
\begin{aligned}
	& \Delta X_{t+i}^{h}=\Delta X_{t}^{h} \\
	&\ \ \ \ \ \ \ \ \ \  +\sum\limits_{j=0}^{i-1}(f_{X_{t-1}^{h}}^{h}{}')^{j}(f_{X_{t-1}^{h}}^{h}{}'\Delta X_{t}^{h}+O({{\left\| \Delta X_{t}^{h} \right\|}^{2}})) \\
	&\ \ \ \ \ \ \ \ \ \  +\sum\limits_{j=0}^{i-1}(f_{X_{t-1}^{h}}^{h}{}')^{j}(f_{U_{t-1}^{h}}^{h}{}'\Delta U_{t+j}^{h}+O({{\left\| \Delta U_{t+j}^{h} \right\|}^{2}})) \\
	& \Delta Y_{t+i}^{h}={{C}^{h}}\Delta X_{t+i}^{h} \\ 
\end{aligned}
\end{equation}

\begin{figure}[!t]
	\centering
	\includegraphics[width=3.2in]{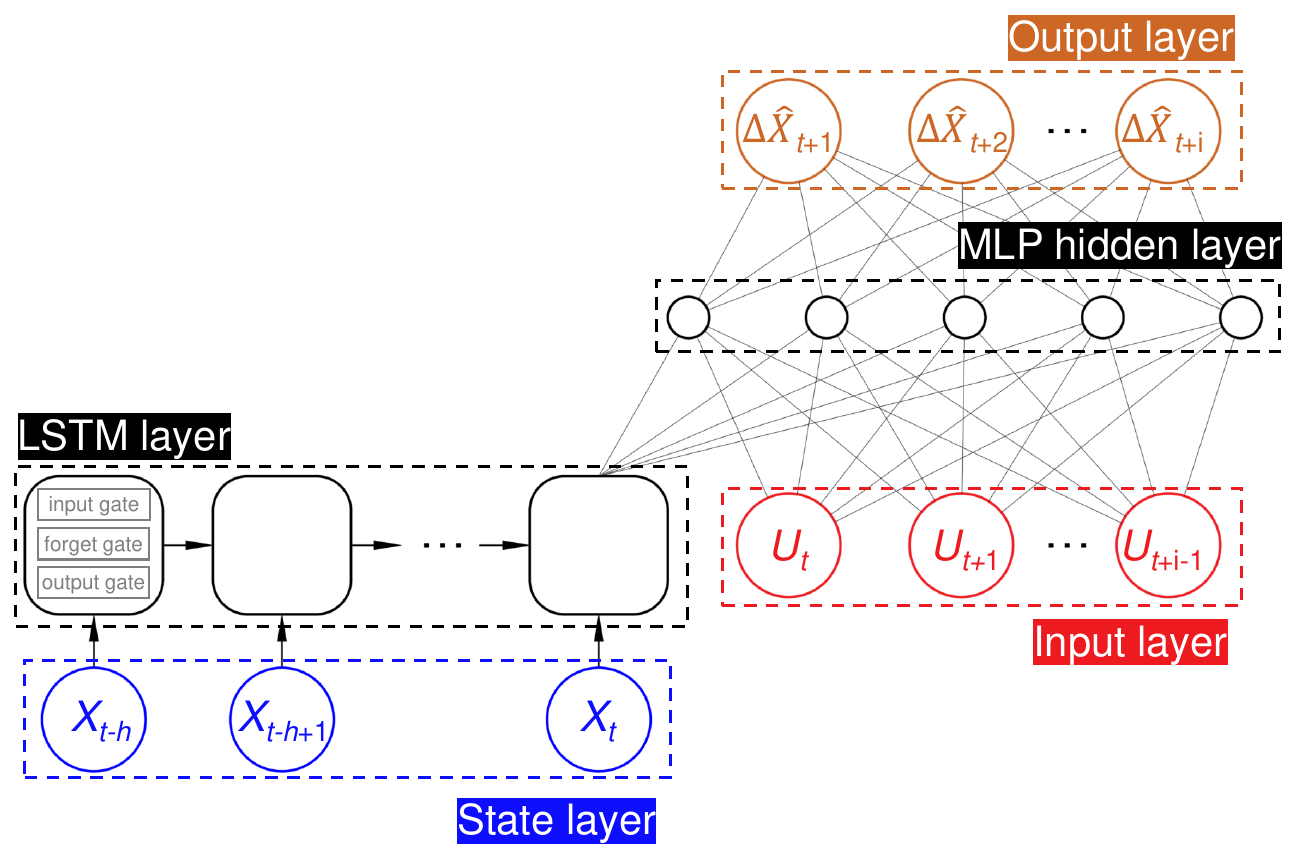}
	\caption{Offline SSMP model. The LSTM includes input gates, forget gates, and output gates, with its input being the system's state. The input for MLP is the system's control signals, which must be transformed into a one-dimensional format if they are originally multidimensional.}
	\label{LSTM_MLP}
\end{figure}

As delineated in (\ref{eq3}), the future state primarily comprises free and forced responses \cite{jordanou2021echo},\cite{plucenio2007practical}. The free response is only associated with the states $X_{t}^{h}$ and their partial derivatives, while the forced response is not only related to the historical states $X_{t}^{h}$, historical inputs $U_{t-1}^{h}$, and their partial derivatives, but is also influenced by the future input signal $\left[ \begin{matrix}
	\Delta U_{t}^{h} & \Delta U_{t+1}^{h} & \ldots  & \Delta U_{t+i-1}^{h}  \\
\end{matrix} \right]$. This indicates that we should extract partial derivative relationships from past data and combine them with upcoming input signals to infer future states. Since ${{U}_{t-1}}$ represents the already acquired historical inputs, utilizing ${{U}_{t+i}}={{U}_{t-1}}+\Delta {{U}_{t+i}}$ directly in place of $\Delta {{U}_{t+i}}$ can achieve similar fitting results and may facilitate a more effective optimization of the cost function. Similarly, use ${{X}_{t+i}}={{X}_{t-1}}+\Delta {{X}_{t+i}}$ to replace $\Delta {{X}_{t}}$ and the following predictive model can be designed: 
\begin{equation}\label{eq4}
\begin{aligned}
	& {{\hat{X}}_{t+1:t+i}}=G(F(X_{t}^{h},U_{t-1}^{h};{{w}_{F}},{{b}_{F}}),{{U}_{t:t+i-1}};{{w}_{G}},{{b}_{G}}) \\	
	& \ \ \ \ \ \ \ \ \ \ \ \ \ + {{X}_{t-1}} \\
\end{aligned}
\end{equation}
where $w$ and $b$ represent the weights and biases in the neural network respectively. The offline neural network predictive model is illustrated in Fig.\ref{LSTM_MLP}. We utilize LSTM to model the nonlinear equation $F$. LSTM features specialized memory cells that improve its ability to retain information from previous states and inputs. These cells are particularly effective at modeling complex, high-order systems that may exhibit hysteresis and delays, leading to strong generalizations in predictions. Since the dimensions of the input signal $U$ differ from the state dimensions, incorporating an encoder with LSTM would significantly increase computational demands. Therefore, we select an MLP as the predictive model $G$, simplifying gradient computation and optimizing performance. The input to model $G$ consists of the output from the LSTM unit along with the future input signal $U$. When using the model (\ref{eq4}), it is essential to establish a predictive horizon. This horizon can be adjusted through rolling iterations or by modifying the output of the learning model.

It is important to emphasize that we do not use the network to predict future states directly; instead, we focus on predicting the changes in states relative to the current moment. This approach offers several advantages because learning-based methods are inherently prone to prediction errors. When we attempt to make direct predictions about future states, the network's output may fluctuate around the actual values, potentially leading to significant errors that can cause predictions to stray from the original trend of state changes. On the other hand, predicting changes in states can help minimize fitting errors and improve the accuracy of trend predictions in state changes.

\subsection{Data Collection}\label{Methodology-B}
The quality of data collection is essential for the learning model's predictive capabilities. We aim to obtain the sensitivity of the system's states to the input $U$ in the vicinity of each state point. The ideal data dimension should be as large as the $(n + m)$ state-input space of the system and the data range should be uniformly distributed across all accessible dimensions. However, collecting data this way is very labor-intensive, and achieving complete coverage of all dimensions presents significant challenges.

Data collection strategies are generally classified into open-loop and closed-loop methods. The open-loop approach gathers system response data by applying random inputs. While this method is straightforward, it can raise safety concerns and often fails to adequately cover the entire state space. On the other hand, closed-loop techniques utilize predesigned controllers and trajectories, adjusting the trajectory profiles to cover each state point effectively. Although this approach minimizes the risks associated with operating outside of safe limits, it presents challenges. The data collected may still exhibit distributional bias in the multi-dimensional input space, which can result in skewed insights \cite{brunke2022safe}.

\begin{figure}[!t]
	\centering
	\includegraphics[width=3.5in]{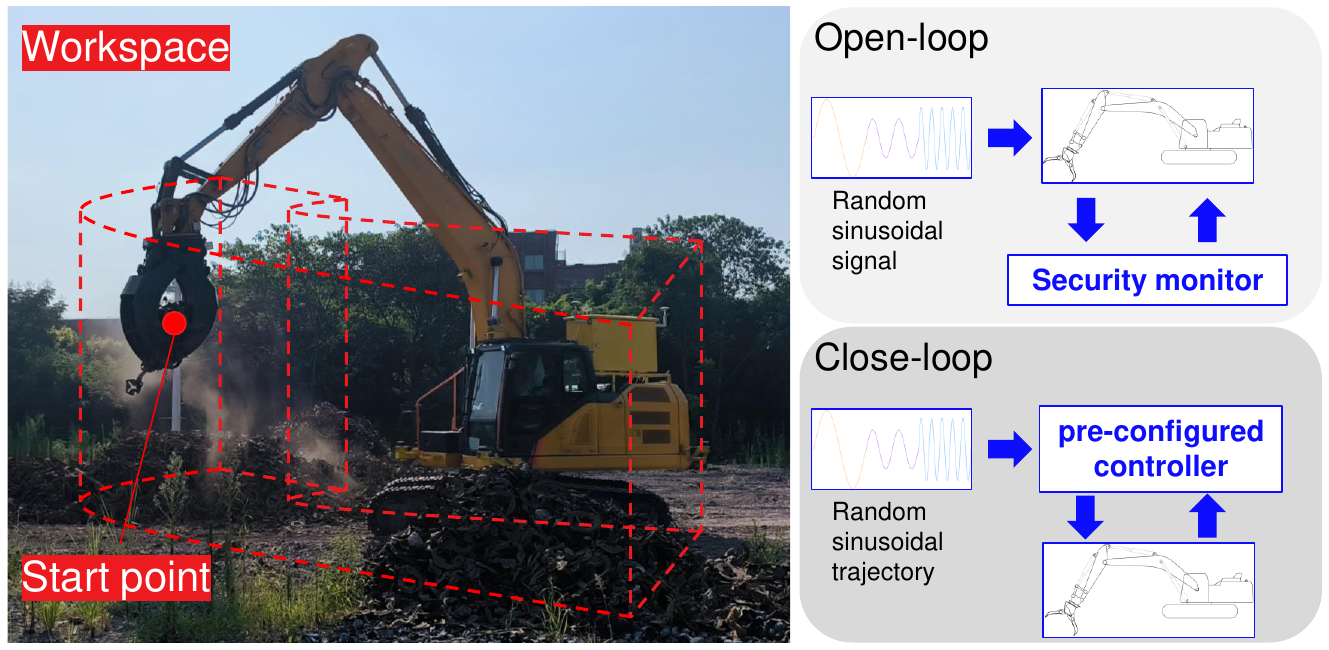}
	\caption{Data Collection Method. The workspace is a subset of the robot's reachable space. The safety monitor calculates the joint cartesian coordinates and determines whether it exceeds the workspace. The pre-configured controller is typically the controller provided by the manufacturer. }
	\label{data_collection}
\end{figure}

As illustrated in Fig.\ref{data_collection}, we have integrated the two previously mentioned approaches and established safety protocols for data collection. Since robots exhibit similar motion patterns when performing the same tasks, we determine their maximum operational workspace based on these tasks. We then gather data within this defined space using a combination of open-loop and closed-loop control strategies. This approach ensures that data collection is efficient and secure, leveraging the consistent motion characteristics observed in robots executing similar tasks. During the open-loop data collection phase, we introduce a sine input signal with a random magnitude and frequency within a specified range. The system continuously monitors its state to ensure it remains within the safe workspace. If any predefined threshold is exceeded, the input signal is promptly halted, and the system resets to the starting point. In the closed-loop collection phase, a preset sine trajectory is defined with random magnitude and frequency that fits within the operational workspace. This trajectory is tracked using a pre-configured controller, ensuring that the data collection process meets established safety and operational constraints.

\subsection{Online Learning Model for Variations}\label{Methodology-C}
\begin{figure*}[!t]
	\centering
	\includegraphics[width=6.4in]{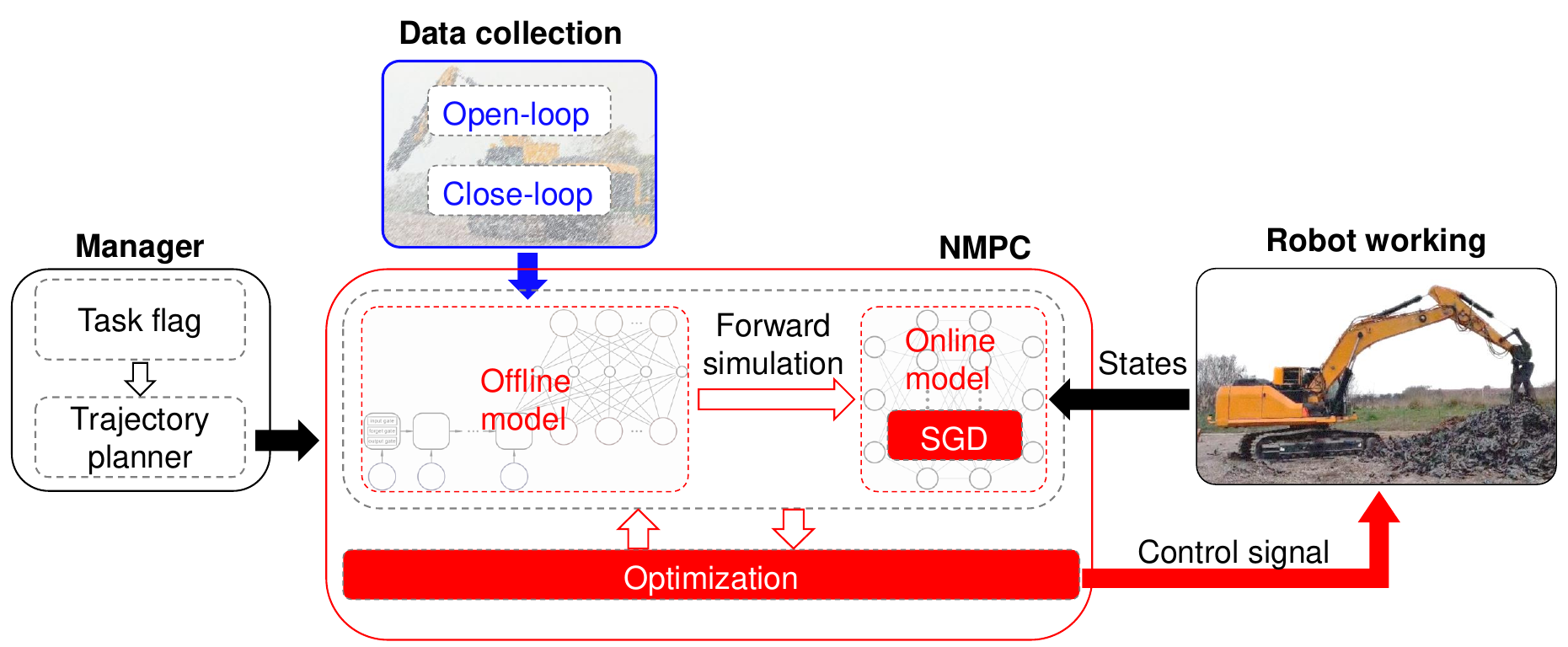}%
	\hfil
	\caption{Block diagram representation of the NMPC. After the manager issues a task, trajectory planning and online model resetting are initiated concurrently. The red block contains the main parts of NMPC, including the offline model, online model, online learning, and online optimization. Solid arrows represent data exchange between different components, while hollow arrows indicate data transfer within NMPC.}
	\label{NMPC}
\end{figure*}
Robots frequently interact with their external environment while performing tasks such as transportation and grasping. These interactions can cause changes in the system's internal parameters, which may reduce the accuracy of the offline-trained predictive model. A decline in model accuracy can adversely impact the control performance of NMPC. Therefore, it is crucial to account for these interaction dynamics to ensure the effectiveness and reliability of robotic control systems across various operational contexts. In response to the variations, traditional methods often rely on sensors to measure external parameters or treat them as disturbances, adjusting inputs accordingly to minimize their effects. However, in many engineering situations, inadequate sensors can hinder robots from accurately detecting the precise impacts of their interactions with the external environment. Drawing inspiration from free and forced response processes, we view the offline-trained model as representing the system's nominal behavior without external interference. In contrast, interactive processes introduce unknown disturbances, leading to a forced response from the robots. This perspective allows robots to adapt to unexpected changes and maintain optimal performance. As a result, we have developed an online learning approach designed to detect and accommodate variations. This approach combines an online learning neural network model with the existing offline-trained model, aiming to enhance the accuracy of model predictions in real time.

We initially train a basic offline model without external interaction. During the task, when the robot faces environmental changes or experiences varying loads, the accuracy of the offline model becomes compromised. Then, we record the actual state $X$ and the forward simulation $X_hat$ of the offline model throughout the task execution process and define the mismatch as ${{X}_{error}}=X-\hat{X}$ which can reflect the changes in the current system model relative to the no-load model. We employ an MLP to facilitate online training $H$, as follows:
\begin{equation}\label{eq5}
	\begin{aligned}
		& {{\hat{X}}_{error}}{{\ }_{t+1:t+i}}=H(X_{t}^{h},U_{t-1}^{h},{{U}_{t:t+i-1}};{{w}_{H}},{{b}_{H}}) \\
	\end{aligned}
\end{equation}

Compared to the serial structure of LSTM, training an MLP is less computationally intensive, which accelerates the convergence process. This advantage makes the MLP particularly useful in situations requiring quick model updates. The online model maintains the same input and output specifications as the offline model, ensuring continuity and ease of implementation. The training dataset is generated from historical data collected during the system's online operations, enabling the model to learn from past experiences and improve its predictions. For instance, we calculate the error between the historical actual state $\left[ \begin{matrix}
	{{X}_{t-h}} & {{X}_{t-h+1}} & \ldots  & {{X}_{t}}  \\
\end{matrix} \right]$ and the prediction state $\left[ \begin{matrix}
{{{\hat{X}}}_{t-h}} & {{{\hat{X}}}_{t-h+1}} & \ldots  & {{{\hat{X}}}_{t}}  \\
\end{matrix} \right]$ from the offline model at time $t$. After obtaining the error, we employ stochastic gradient descent (SGD) to iteratively refine the weights $w$ and biases $b$ of the MLP in an online setting to minimize the error. The formula is as follows:
\begin{equation}\label{eq6}
	\begin{aligned}
		& M=\left\| X_{t}^{h}-\hat{X}_{t}^{h}-{{{\hat{X}}}_{error}}\ _{t}^{h} \right\|_{2}^{2} \\
		& {{w}_{H}}(k)={{w}_{H}}(k-1)-\frac{{{\eta }_{w}}}{B}\sum\limits_{i=k-B}^{k}{\frac{\partial M}{\partial U}} \\ 
		& {{b}_{H}}(k)={{b}_{H}}(k-1)-\frac{{{\eta }_{b}}}{B}\sum\limits_{i=k-B}^{k}{\frac{\partial M}{\partial U}} \\ 
	\end{aligned}
\end{equation}
where $k$ is the current iteration and $B$ is batch size. Learning rate ${{\eta }_{w}}$ and ${{\eta }_{b}}$ are hyperparameters that determine the adaptability of the online model. Starting with small initial weights and biases is essential for every online learning process. This cautious approach helps the model avoid the interference of errors caused by random initial parameters. Employing mini-batch training enhances the stability of the training process and improves the generalization of the results.

\subsection{Data-Driven NMPC}\label{Methodology-D}
Industrial robots often function as complex, MIMO-coupled nonlinear systems. Their controllers are designed to meet the demands of production and manufacturing tasks. We also aim to enhance the robots' energy efficiency, as this can result in significant economic savings, which is crucial for the economic viability and sustainability of industrial automation. NMPC is particularly effective in control systems that involve multi-objective optimization. It utilizes a specified nonlinear system dynamics model to minimize a cost function, denoted as $J$, which helps extract the most effective sequence of control inputs. This cost function accounts for the system's inherent constraints and desired behaviors. Choosing an appropriate cost function is crucial, as it balances operational performance and energy consumption. In the NMPC framework shown in Fig.\ref{NMPC}, with a predictive horizon set to $N$, the optimization problem can be described as:
\[
{{U}_{t:t+N-1}}=\arg {{\min }_{U}}\sum\limits_{i=1}^{N}{J({{R}_{t+i}},{{X}_{t+i}},{{U}_{t+i-1}})}
\]
\begin{equation}\label{eq7}
	\begin{aligned}
		& s.t.\ \ {{X}_{t+1:t+N}}=G(F(X_{t}^{h},U_{t-1}^{h}),{{U}_{t:t+N-1}}) \\
		& \ \ \ \ \ \ \ \ \ \ \ \ \ \ \ \ \ \ \ \ +H(X_{t}^{h},U_{t-1}^{h},{{U}_{t:t+N-1}}) \\
		& \ \ \ \ \ \ \ \ \ \ \ \ \ \ \ \ \ \ \ \ +{{X}_{t-1}} \\ 
		& \ \ \ \ \ \ S({{X}_{t+1:t+N}},{{U}_{t:t+N-1}})\le 0 \\ 
	\end{aligned}
\end{equation}
where ${{R}_{t+i}}$ represents the reference state at time $t+i$. System dynamics are defined by the combined behavior of offline and online models. $S$ includes the system's constraints, such as the robot's operational workspace and the input signal saturation. The cost function is generally convex and incorporates task performance and energy consumption factors. 

Using a nonlinear learning model complicates the direct application of linear MPC. Although linearizing the nonlinear model could be a potential solution, it may decrease predictive accuracy. To maintain the reliability of predictions, we choose the GD method to address the optimization challenges associated with the nonlinear model. Although this way may not always reach the global optimum in a finite number of iterations, they consistently yield satisfactory control outcomes by optimizing each time step. This method is flexible, not limited to a single cost function, and can adapt to changes in the desired behavior of the system, allowing for continuous optimization. The interior-point method integrates constraints as penalty terms within the cost function, effectively addressing the limitations of a system. This approach is particularly beneficial for linear programming and non-linear convex optimization problems. Alternatively, a saturation function can be directly designed to prevent the system from exceeding its operational limits. The predictive learning model proposed in this work facilitates efficient derivative analysis and computation, making implementing GD methods easier. This approach effectively circumvents the complex computational requirements of finite difference methods, significantly reducing computation time. For example, the conventional GD optimization process begins with all variable initial values set to zero:
\begin{equation}\label{eq8}
	\begin{aligned}
		{{U}_{t:t+N-1}}(k)={{U}_{t:t+N-1}}(k-1)-{{\eta }_{U}}\frac{\partial 	J(k-1)}{\partial {{U}_{t:t+N-1}}(k-1)}
	\end{aligned}
\end{equation}

In the process of taking the partial derivative of $J$ concerning $U$, $\frac{\partial G}{\partial U}$ and $\frac{\partial H}{\partial U}$ will be calculated. Since $G$ and $H$ are designed as MLPs, their derivatives can be directly computed using the chain rule. Taking $G$ as an example, assuming it is an MLP with two hidden layers, the forward computation and derivative formulas are shown as follows:
\begin{equation}\label{eq9}
	\begin{aligned}
		& {}^{1}{{V}_{G}}={}^{1}{{w}_{G}}\cdot [\begin{matrix}F 
		& {{U}_{t:t+N-1}}  \\
		\end{matrix}]+{}^{1}{{b}_{G}} \\ 
		& {}^{2}{{V}_{G}}={}^{2}{{w}_{G}}\cdot \sigma ({}^{1}{{V}_{G}})+{}^{2}{{b}_{G}} \\ 
		& G={}^{3}{{w}_{G}}\cdot \sigma ({}^{2}{{V}_{G}})+{}^{3}{{b}_{G}} \\ 
		& \frac{\partial G}{\partial {{U}_{t:t+N-1}}}={}^{3}{{w}_{G}}\cdot \sigma '({}^{2}{{V}_{G}})\cdot {}^{2}{{w}_{G}}\cdot \sigma '({}^{1}{{V}_{G}})\cdot {}^{1}{{w}_{G}} \\ 
	\end{aligned}
\end{equation}
where $\sigma$ represents the activation function, ${}^{i}{{w}_{G}}$, ${}^{i}{{b}_{G}}$, and ${}^{i}{{V}_{G}}$ are the weights, biases, and outputs of MLP’s $i$ layer respectively. Only the part related to $U$ must be taken when computing the partial derivatives.

The criteria for terminating an iterative process typically include three main factors: reaching a maximum number of iterations, achieving a target value for the loss function, and observing a change in the loss function below a specified threshold. In our case, we have set the maximum number of iterations as the termination condition. This decision allows us to manage computational resources while optimizing the process effectively. Given the inherent variability in the cost function, relying solely on a fixed target value or the magnitude of change in the loss function may not be sufficiently flexible.

The overall NMPC algorithm works as Algorithm \ref{NMPC_algorithm}: When the robot starts a task, it continuously gathers the filtered current state from its sensors. Then, the error in the predictive model is calculated, which is used to improve the online model. Meanwhile, the controller optimizes the best control sequence based on the updated predictive model.

\begin{algorithm}[t]
	\caption{NMPC}\label{NMPC_algorithm}
	\begin{algorithmic}[1]
		\STATE Train the offline learned model based on motion data;
		\STATE Initial the online learning model;
		\STATE \textbf{While} task \textbf{do}
		\STATE \hspace{0.5cm} \textbf{Input:} System state $X$;
		\STATE \hspace{0.5cm} \textbf{For} $i\ in\ [1:k_1]$ \textbf{do}
		\STATE \hspace{1.0cm} Set the $U=0$, iteration = $i$
		\STATE \hspace{1.0cm} Predict the system output $\hat{Y}$
		\STATE \hspace{1.0cm} Calculate jacobian matrix
		\STATE \hspace{1.0cm} Update control signal through GD
		\STATE \hspace{0.5cm} Record the input and output data;
		\STATE \hspace{0.5cm} \textbf{For} $j\ in\ [1:k_2]$ \textbf{do}
		\STATE \hspace{1.0cm} Calculate the error of prediction
		\STATE \hspace{1.0cm} Error backpropagation and update online model
		\STATE \hspace{0.5cm} \textbf{Output:} Control signal $U$;
	\end{algorithmic}
\end{algorithm}

\section{Experimental Setup}\label{Experimental Setup}
Through NMPC, we aim to achieve a balance between industrial robots' work performance and energy efficiency to reduce production and manufacturing costs. To assess the efficacy of our proposed approach, we carry out simulations and practical experiments with a hydraulic excavator. In the following, we will provide a detailed introduction to experiments and simulations involving the hydraulic excavator.
\subsection{Hydraulic Excavator}\label{Experimental Setup-A}
\begin{figure*}[!t]
	\centering
	\includegraphics[width=6.4in]{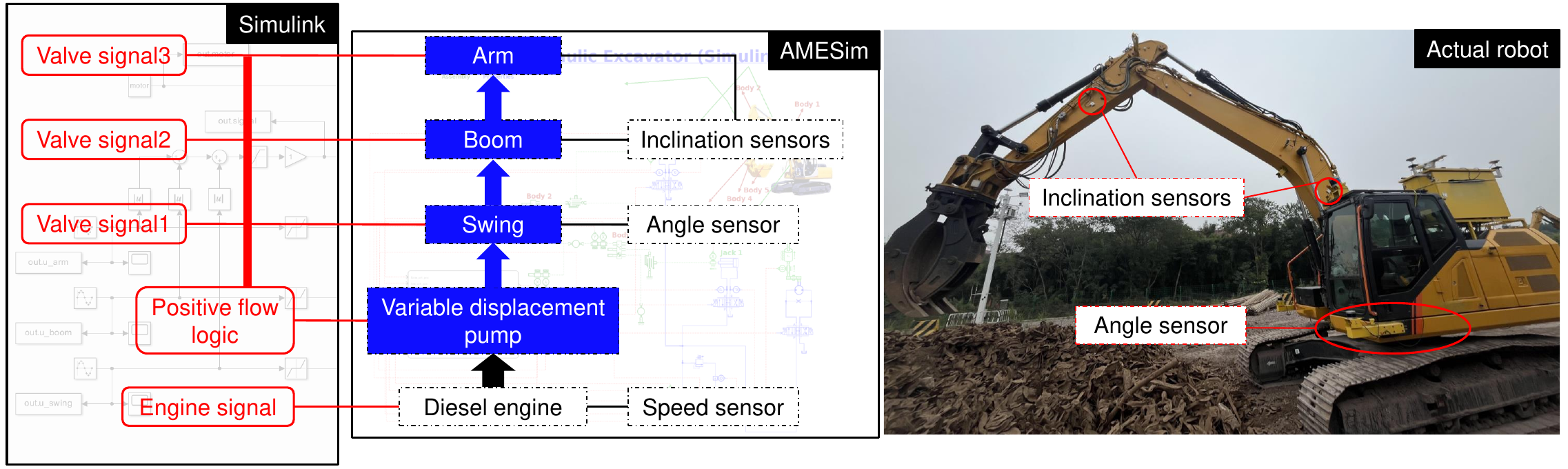}%
	\hfil
	\caption{The experimental machine and corresponding simulation model. The simulation model is constructed based on AMESim demo and using the same sensors as the actual robot. We proactively introduce nonlinearities such as dead zones, leakages, and friction, and optimize the control signals in real-time within Simulink s-function.}
	\label{Simulation}
\end{figure*}

Our analysis focuses on a specific model of a 22-ton hydraulic excavator. To enhance our understanding of its operational dynamics, we have developed a simplified simulation model targeting the manipulators using AMESim and Simulink. As illustrated in Fig.\ref{Simulation}, we employ a variable-displacement piston pump to supply hydraulic fluid to three actuators, based on a hydraulic excavator demonstration in AMESim, effectively replicating the oil supply circuitry. Furthermore, we consider a positive flow system, characterized by control signals that simultaneously influence multiple solenoid proportional valves and the variable displacement pump. This system manages both the opening of the proportional valves and the pump's output flow rate, closely resembling the operation of the actual machine. We also incorporate common nonlinear characteristics, such as dead zones and leakage, into our model. The control targets include the engine speed and the electromagnetic proportional valves for the three primary actuators of the hydraulic excavator. In the actual machine, the engine is equipped with ten forward gears. To accommodate various work tasks, we select the third, fifth, and seventh gears to represent the simulation model's low, medium, and high operational ranges. We have omitted the inertia related to adjusting the engine's speed because gear shifts can be performed at will and are completed almost instantaneously in the actual machinery. Additionally, a relief valve is connected to the pump's outlet, serving as a safety feature, and the overflow can be used to determine wasted energy.

The experiment is conducted on a 22-ton hydraulic excavator after the simulation. We have installed inclination sensors on the boom and arm linkages to acquire joint angle data, and the swing angle can be measured through the built-in angle sensor. To enable remote control functionality, we use two laptops, referred to as "transmission" and "control." The user datagram protocol (UDP) is designed to meet our communication needs, which allows for a communication frequency of 50 Hz. The transmission computer is situated in the cab of the excavator and serves the purpose of relaying position, speed, and other relevant data back to the control computer. It also transmits control commands to the drivers. Meanwhile, the "control" computer is responsible for real-time task planning and calculating control commands based on the state data it receives.

\subsection{Offline and Online Learning Model}\label{Experimental Setup-B}
We select $\left[ \begin{matrix} q & {\dot{q}} & {\ddot{q}}  \\\end{matrix} \right]$ of the swing, boom, and arm joints as the state vector $X\in {{\mathbb{R}}^{9}}$ and input signal $U=\left[ \begin{matrix}
{{u}_{swing}} & {{u}_{boom}} & {{u}_{arm}} & {{\omega }_{engine}}  \\ \end{matrix} \right]\in {{\mathbb{R}}^{4}}$. This is because obtaining pressure data can be highly costly for standard hydraulic excavators that have not been modified. In contrast, joint angle signals, essential for monitoring the machine's movement and condition, can be easily captured using inclination sensors. This approach is cost-effective and applicable across all existing machines, regardless of whether they have been retrofitted with advanced sensors. In the offline mode, the LSTM is configured as a single layer, with 128 hidden nodes and h denoting the number of cells. The input to each unit within the LSTM layer is a 13×1 tensor that combines the state vector and the input vector, referred to as ${{\left[ \begin{matrix} X & U  \\ \end{matrix} \right]}^{T}}\in {{\mathbb{R}}^{13}}$ The output of the LSTM layer is a 128×1 tensor, which encapsulates the hidden information derived from the network's processing. The MLP includes two hidden layers, each with a size of 128 neurons, and the activation function for each hidden layer is ReLU due to its computational efficiency and faster processing. The input to the MLP is a (128+4$i$)×1 tensor that merges the output from the LSTM with the arranged future input signals ${{U}_{t:t+i-1}}$ and its output is the state change, which is defined as the joint angle change sequence $\left[ \begin{matrix} \Delta {{q}_{swing}} & \Delta {{q}_{boom}} & \Delta {{q}_{arm}}  \\ \end{matrix} \right]_{t:t+i-1}^{T}$. 

The methodology for obtaining offline data is detailed in Section \ref{Methodology-B}. We have collected approximately 1 million and 0.3 million time steps, which amounts to about 14 hours and 4.2 hours of data for simulations and experiments, respectively. Before training, we apply min-max normalization to both the input and output datasets, ensuring that all values are scaled to the interval (0, 1). This preprocessing step is essential for placing the data on a comparable scale, which promotes more effective learning processes. The training is conducted using a supervised method, with the forward loss calculated as $L=\left\| {{X}_{t+1:t+n}}-{{{\hat{X}}}_{t+1:t+n}} \right\|_{2}^{2}$, and the weights of the LSTM and MLP updated using the adaptive moment estimation method (Adam). The training process has been run for 400,000 iterations, with a batch size of 256 samples and a learning rate of 0.001 using PyTorch.

The online learning model features two hidden layers, each composed of 64 neurons, utilizing the ReLU as the activation function. The input is generated by expanding all inputs from the offline model into a tensor referred to as (13$h$+4$i$)×1, and the output is the same as the output of the offline model. The training method is described in Section \ref{Methodology-C}. The number of iterations per NMPC cycle is set to 30 to reduce the computational demands during real-time operations.

\subsection{Cost Function and Optimization}\label{Experimental Setup-C}
This work addresses a classic engineering task involving the handling of scrap steel. Specifically, the excavator grasps scrap steel from one location and transfers it to another, typically a truck. This task provides a practical scenario for research, highlighting the common challenges and operations excavators encounter in industrial environments. We can also evaluate the efficacy of our methods. This task requires the excavator to reach specified target points precisely. The operational efficiency of our system should meet or exceed that of an average operator while simultaneously striving to minimize fuel consumption to the lowest possible levels. This dual emphasis on efficiency and cost-effectiveness is vital for optimizing excavator performance. Consequently, the cost function is designed as follows:

\begin{equation}\label{eq10}
	\begin{aligned}
		& J={{({{R}_{Y}}{{}_{t+1:t+N}}-{{{\hat{Y}}}_{t+1:t+N}})}^{T}}a({{R}_{Y}}{{}_{t+1:t+N}}-{{{\hat{Y}}}_{t+1:t+N}}) \\
		&\ \ \ \ +{{({{R}_{\dot{Y}}}{{}_{t+1:t+N}}-{{{\hat{\dot{Y}}}}_{t+1:t+N}})}^{T}}b({{R}_{\dot{Y}}}{{}_{t+1:t+N}}-{{{\hat{\dot{Y}}}}_{t+1:t+N}}) \\ 
		&\ \ \ \ +{{\omega }_{engine}}{{}_{t+1:t+N}}^{T}c{{\omega }_{engine}}{{}_{t+1:t+N}} \\ 
	\end{aligned}
\end{equation}
subject to
\begin{equation}\label{eq11}
	\begin{aligned}
		& {{{\hat{Y}}}_{t+1:t+N}}=G(F(X_{t}^{h},U_{t-1}^{h}),{{U}_{t:t+N-1}}) \\
		&\ \ \ \ \ \ \ \ \ \ \ \ \ \ +H(X_{t}^{h},U_{t-1}^{h},{{U}_{t:t+N-1}})+{{Y}_{t-1}} \\ 
		& {{{\hat{\dot{Y}}}}_{t+1:t+N}}=\frac{{{{\hat{Y}}}_{t+1:t+N}}-{{{\hat{Y}}}_{t:t+N-1}}}{\Delta t} \\ 
		& U\in [\begin{matrix}
			{{U}_{\min }} & {{U}_{\max }}  \\ 
			\end{matrix}] \\ 
		& {{\omega }_{engine}}\in \left\{ \begin{matrix}
			low & medium & high  \\
			\end{matrix} \right\} \\ 
		& {{t}_{switch}}>1s \\ 
			\end{aligned}
\end{equation}
where ${R}_{Y}$ and ${R}_{\dot{Y}}$ respectively represent the reference position and velocity. $a$, $b$, and $c$ the weight coefficients. Here, we consider both position errors and velocity errors. This approach facilitates smooth motion control by effectively managing velocity, reflecting the proposed method's strengths in addressing multi-objective optimization challenges. Furthermore, when position errors are minimal, the fitting error in the neural network prediction model may lead to the generation of slight over-regulation signals. Introducing the velocity error part can reduce vibrations, enhancing the system's stability. Additionally, the engine's rotational speed is incorporated into the cost function to balance control performance and fuel efficiency. The engine speed can only be selected from three predefined levels: low, medium, and high. Since diesel engine gear shifting takes time, we set a ${t}_{switch}$ that represents the cooldown period following a change in engine speed, ensuring that a minimum of one second elapses after each gear shift before another adjustment can be made.

Optimization aims to identify the most effective input sequence ${{U}_{t:t+N-1}}$, which will minimize the cost function $J$. To lessen the impact of predictive errors, we have developed an enhanced method for parameter adaptation as (\ref{eq12}). Before each iteration of NMPC, the learning rate must be adjusted based on the current position error. This adaptive approach ensures that when the manipulators are significantly distant from the reference position, the initial learning rate is employed to facilitate substantial adjustments. Conversely, when the positional error is minimal, the learning rate is appropriately reduced, thereby averting the risk of over-regulation signals. Conversely, when the positional error is minimal, the learning rate is decreased appropriately to prevent the risk of over-regulation signals. By implementing such dynamic adjustments to the learning rate, we can ensure that our model responds sensitively to variations in error magnitude. 
\begin{equation}\label{eq12}
{{\eta }_{U}}=\left\{ \begin{aligned}
	& {{\eta }_{U}},\left| {{R}_{Y}}-Y \right|\ge e \\ 
	& {{\eta }_{U}}\left| {{R}_{Y}}-Y \right|,\left| {{R}_{Y}}-Y \right|<e \\ 
\end{aligned} \right.
\end{equation}
where $e$ is the preset error threshold. It is important to note that we utilize the average of the entire sequence instead of the first element as the control signal. In contrast to single-step forecasting, the proposed multi-step predictive model deals with sequences for both inputs and outputs. The optimization result is affected by all elements within the sequence, necessitating a value that considers the collective influence of these elements. The example (detailed in Section \ref{Results and Discussions-C}) demonstrates that utilizing the control sequence's mean value yields superior control performance compared to relying solely on the first signal.

\subsection{Metrics}\label{Experimental Setup-D}
The evaluation of model prediction ability is conducted utilizing Root Mean Square Error (RMSE) and Average RMSE (ARMSE ), as shown in (\ref{eq13}). The RMSE is used as a measure of the error between the actual state $Y$ and the predicted state $Y_hat$ within the predictive horizon. 
\begin{equation}\label{eq13}
\begin{aligned}
		& \text{RMSE(}Y,\hat{Y}\text{)=}\sqrt{\frac{1}{N}\left\| {{Y}_{t+1:t+N}}-{{{\hat{Y}}}_{t+1:t+N}} \right\|_{2}^{2}} \\ 
		& \text{ARMSE(}{{Y}_{t}},{{{\hat{Y}}}_{t}}\text{)=}\frac{1}{t}\sum\limits_{i=0}^{t}{\text{RMSE(}{{Y}_{i}},{{{\hat{Y}}}_{i}}\text{)}} \\ 
\end{aligned}
\end{equation}

Regarding energy efficiency, we establish a metric $E$ for the energy utilization efficiency, which is detailed in (\ref{eq14}):
\begin{equation}\label{eq14}
	\begin{aligned}
		{{E}}(t)=1-\frac{\int_{0}^{t}{{{Q}_{overflow}}dt}}{\int_{0}^{t}{{{\omega }_{engine}}\cdot {{L}_{pump}}dt}}
	\end{aligned}
\end{equation}
where ${{L}_{pump}}$ is the maximum nominal displacement of the oil pump, and ${{Q}_{overflow}}$ is the flow rate through the relief valve.

\section{Results and Discussions}\label{Results and Discussions}
In this section, we will demonstrate the advantages of the method proposed in this work based on the simulation environment in terms of predictive modeling and control capabilities. Finally, we have experimentally verified the method on a 22-ton hydraulic excavator.

\subsection{Offline Model and Characteristic Analysis}\label{Results and Discussions-A}
\begin{figure}[!t]
	\centering
	\includegraphics[width=3.5in]{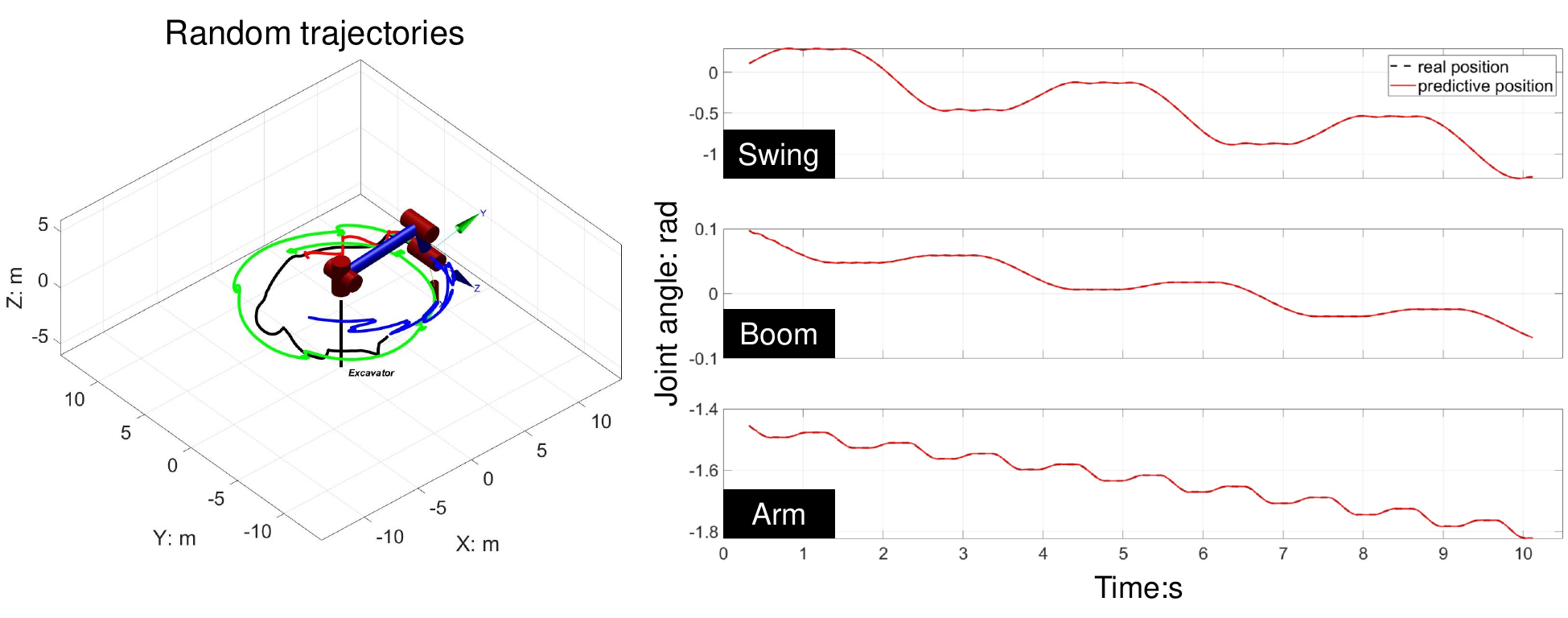}
	\caption{The predictions of trajectories under random sine input signal. The left diagram employs the MATLAB robotic toolbox to depict the trajectories of the end-joint, each distinguished by a unique color and the right diagram shows one of the examples.}
	\label{trajectory}
\end{figure}

The comparative baseline method is \cite{wang2020deep}, which is the most commonly used multi-step prediction neural network model, and both methods utilize the same dataset. We provide random sinusoidal input signals to generate the motion trajectory of the hydraulic manipulators, as shown in Fig.\ref{trajectory}. The model's predictive precision has been evaluated through various random trajectories based on historical length $h=20$ and predictive horizon $N=10$. The results are detailed in Table \ref{ARMSE of Prediction}. Although the networks used the same dataset, the proposed method's predictive accuracy consistently exceeds that of the baseline, demonstrating excellent robustness. This significant performance improvement is mainly due to the LSTM's ability to extract relevant historical data information effectively.

\begin{table}[t]
	\begin{center}
	\caption{ARMSE of Prediction $[rad]$}\label{ARMSE of Prediction}
	\setlength{\tabcolsep}{1mm}
	\begin{tabular}{*{7}{c}}
		\toprule
		\multirow{2}*{Trajectory} & \multicolumn{2}{c}{${q}_{swing}$} & \multicolumn{2}{c}{${q}_{boom}$} & \multicolumn{2}{c}{${q}_{arm}$}  \\
		\cmidrule(lr){2-3}\cmidrule(lr){4-5}\cmidrule(lr){6-7}
					& DBN \cite{wang2020deep} 	& Ours 				& DBN \cite{wang2020deep}  	& Ours 				& DBN \cite{wang2020deep}  	& Ours \\
		\midrule
		$Low1$ 		& 2.96e-4 	& \textbf{2.34e-4} 	& 1.38e-4	& \textbf{5.67e-5} 	& 3.04e-4 	& \textbf{1.79e-4}  \\
		$Low2$ 		& 4.61e-4 	& \textbf{2.74e-4} 	& 1.54e-4	& \textbf{3.83e-5} 	& 1.32e-3 	& \textbf{2.99e-4}  \\s
		$Medium1$ 	& 3.54e-4	& \textbf{1.75e-4}	& 2.71e-4	& \textbf{7.47e-5} 	& 1.58e-3 	& \textbf{3.96e-4}  \\
		$Medium2$ 	& 2.98e-4	& \textbf{1.74e-4}	& 1.48e-4	& \textbf{3.36e-5} 	& 1.95e-3 	& \textbf{5.40e-4}  \\
		$High1$ 	& 4.51e-4	& \textbf{2.72e-4}	& 1.33e-4	& \textbf{4.98e-5} 	& 6.33e-4 	& \textbf{1.74e-4}  \\
		$High2$ 	& 4.27e-4	& \textbf{2.57e-4}	& 1.24e-4	& \textbf{4.37e-5} 	& 6.47e-4 	& \textbf{2.31e-4}  \\
	\bottomrule
	\end{tabular}
	\end{center}
\end{table}

Table \ref{ARMSE of Prediction with Different Parameters} presents the ARMSE for a segment of random trajectory under different historical lengths and predictive horizons. The predictive horizon is the most significant factor influencing predictive precision. As the predictive horizon increases, the complexity of the higher-order model information included in the predictive model (\ref{eq4}) also rises, complicating the training process. Extending the historical length within the same predictive horizons can enhance the model's predictive precision. Longer historical data offers the model a richer context, enabling it to make more nuanced predictions by capturing long-term trends. However, this improvement comes with the downside of increased computational complexity, as LSTM models require sequential computations. This trade-off between accuracy and computational efficiency is essential in designing predictive models, particularly when real-time performance or resource limitations are considered. Therefore, it is important to determine appropriate lengths for both historical data and the predictive horizon, taking into account the control plant's nonlinearity.

\begin{table}[t]
	\begin{center}
		\caption{ARMSE of Prediction with Different Parameters $[rad]$}\label{ARMSE of Prediction with Different Parameters}
		\setlength{\tabcolsep}{2.4mm}
		\begin{tabular}{cccc}
			\toprule
			Parameters			& ${q}_{swing}$ & ${q}_{boom}$ 	& ${q}_{arm}$\\
			\midrule
			$h=10, N=5$ 		& 5.42e-5 		& 1.45e-5		& 1.19e-4 \\
			$h=10, N=10$        & 2.83e-4 		& 4.63e-5		& 3.32e-4 \\
			$h=10, N=20$        & 8.81e-4 		& 1.43e-4		& 9.08e-4 \\
			$h=20, N=5$         & 5.09e-5 		& 1.10e-5		& 1.14e-4 \\
			$h=20, N=10$        & 2.48e-4 		& 3.12e-5		& 2.64e-4 \\
			$h=20, N=20$        & 8.47e-4 		& 1.17e-4		& 7.43e-4 \\
			$h=20, N=40$     	& 2.63e-3  		& 3.00e-4		& 2.15e-3 \\
			\bottomrule
		\end{tabular}
	\end{center}
\end{table}

\subsection{Online Learning of External Interactions and Hyperparameters}\label{Results and Discussions-B}
To demonstrate the impact of online learning, we have simulated the scenario of an excavator grasping scrap steel. In this simulation, we apply a substantial load to the terminal joint and utilize random sinusoidal signals to create random motion trajectories for prediction purposes. Table \ref{ARMSE of Hybrid Prediction} shows the predicted ARMSE under different gear power conditions. Upon the introduction of additional load, the predictive precision of the offline model declines, particularly for the boom and arm joints at low gear settings. This decrease is likely due to the increased mechanical stress and altered dynamics experienced by these joints under the influence of the load. In contrast, the swing joint, which has a moment of inertia significantly larger than that of the load, remained relatively unaffected. Integrating the online learning model has led to a notable improvement in predictive accuracy. Across all evaluated datasets, the ARMSE for all joints has been reduced by at least 50\%. This significant enhancement in ARMSE demonstrates that the online learning model can quickly adapt to changes, leading to more reliable predictions.

\begin{figure}[!t]
	\centering
	\includegraphics[width=3.5in]{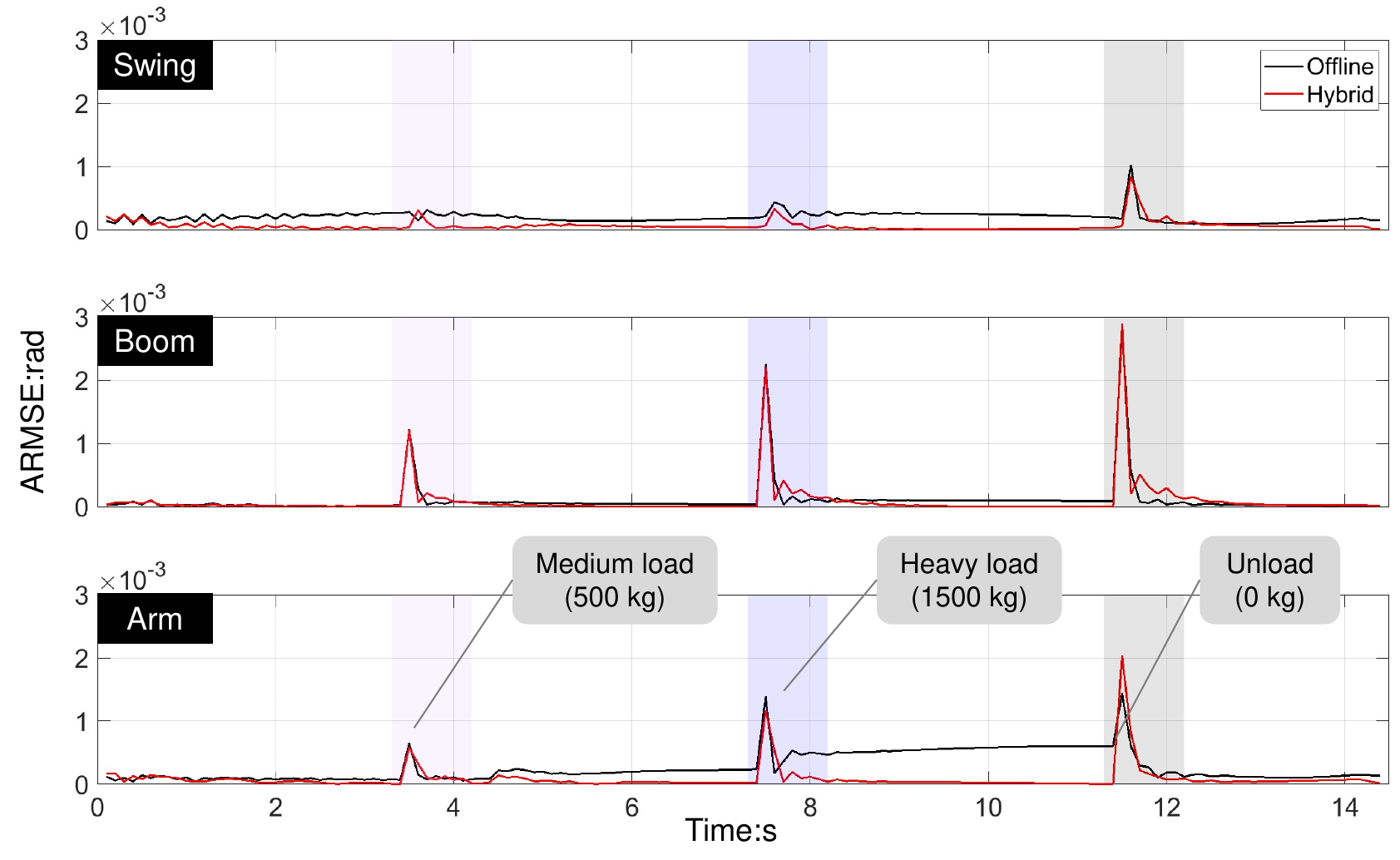}
	\caption{Online learning process. The diagram illustrates the real-time AMRSE, with the additional mass at the end joint representing the robot capturing an object. Hybrid refers to the combination of offline and online model.}
	\label{onlinelearning}
\end{figure}

The process of online learning, when external loads are changed multiple times, is illustrated in Fig.\ref{onlinelearning}. The load is abruptly applied and then removed, similar to a step signal, resulting in a rapid change in the instantaneous prediction error. In response, the online learning model identifies the mismatches in the offline model and adjusts its weights to minimize the error, ensuring that the prediction error progressively decreases and converges over time. The combined offline and online model maintains good predictive performance without affecting the NMPC. Furthermore, the online model builds upon the foundation of the offline model to learn the discrepancies, which results in a comparatively swift convergence rate. The results demonstrate that the proposed online learning approach can address system variations, provide reliable performance across various scenarios, and exhibit generalization ability.

\begin{table}[t]
	\begin{center}
		\caption{ARMSE of Hybrid Prediction $[rad]$}\label{ARMSE of Hybrid Prediction}
		\setlength{\tabcolsep}{1.6mm}
		\begin{tabular}{*{7}{c}}
			\toprule
			\multirow{2}*{Trajectory} & \multicolumn{2}{c}{${q}_{swing}$} & \multicolumn{2}{c}{${q}_{boom}$} & \multicolumn{2}{c}{${q}_{arm}$}  \\
			\cmidrule(lr){2-3}\cmidrule(lr){4-5}\cmidrule(lr){6-7}
						& Offline 	& Hybrid 			& Offline 	& Hybrid			& Offline 	& Hybrid    \\
			\midrule
			$Low$ 		& 2.89e-4	& \textbf{7.93e-5} 	& 1.58e-4	& \textbf{7.20e-5} 	& 1.97e-3	& \textbf{6.47e-4}  \\
			$Medium$ 	& 3.02e-4	& \textbf{1.32e-4}	& 1.43e-4	& \textbf{6.20e-5} 	& 1.83e-3	& \textbf{5.46e-4}  \\
			$High$ 		& 2.09e-4	& \textbf{1.09e-4}	& 1.69e-4	& \textbf{6.99e-5} 	& 9.25e-4	& \textbf{2.87e-4}  \\
			\bottomrule
		\end{tabular}
	\end{center}
\end{table}

It merits emphasis that the input signal significantly impacts the learning process. Our experimental results indicate that when the input signal is monotonous, it substantially enhances the rate of system convergence. This is most common in excavator handling tasks, where simplicity in motion trajectories is directly correlated with improved economic efficiency. Conversely, rapid fluctuations in the input signal may induce oscillation in the manipulators, thereby adversely affecting the convergence rate of the online optimization algorithm. 

The effectiveness of the learning process in an online model primarily depends on the selection of the learning rate and the number of iterations. Misconfiguration of these hyperparameters can negatively affect the model's predictive performance. Table \ref{Percentage Decrease in the ARMSE with Different Hyperparameters} presents the percentage decrease in the ARMSE of the arm joint position compared to the offline model. This metric is a valuable indicator for evaluating the learning performance of the online model. The results show that increasing the learning rate or the number of iterations can improve predictive capability. However, raising both parameters simultaneously may reduce the model's performance, indicating that a balance is necessary to optimize the learning process effectively. For online models with a fixed structure, improvements in predictive performance have an upper limit. In general, increasing the learning rate imposes a lower computational burden than increasing the number of iterations, which enhances the efficiency of real-time processing capabilities.

\begin{table}[t]
	\begin{center}
		\caption{Percentage Decrease in the ARMSE with Different Hyperparameters}\label{Percentage Decrease in the ARMSE with Different Hyperparameters}
		\setlength{\tabcolsep}{2.4mm}
		\begin{tabular}{*{5}{c}}
			\toprule
			\multirow{2}*{Loops} & \multicolumn{4}{c}{Learning rate}\\
			\cmidrule(lr){2-5}
						& 0.001		& 0.005 	& 0.01		& 0.05			\\
			\midrule
			10 			& 27.4\%	& 58.5\%	& 67.7\%	& 68.5\%	\\
			30 			& 47.6\%	& 69.5\%	& 70.0\%	& 62.9\%	\\
			50 			& 58.5\%	& 70.1\%	& 68.7\%	& 62.0\%	\\
			\bottomrule
		\end{tabular}
	\end{center}
\end{table}

\subsection{NMPC and Experimental Validation}\label{Results and Discussions-C}
\begin{figure}[!t]
	\centering
	\includegraphics[width=3.4in]{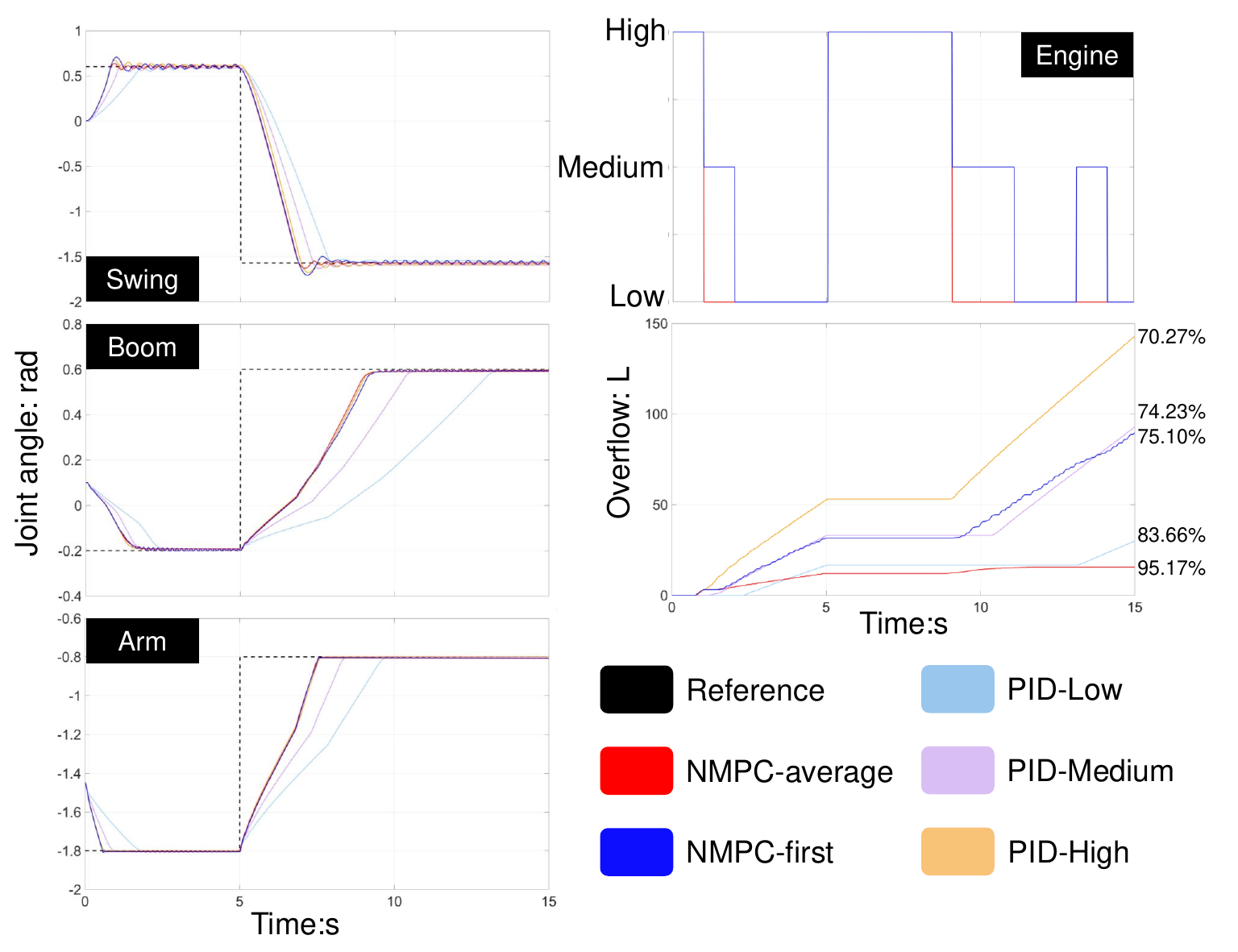}
	\caption{Comparison between the NMPC and a PID controller for the hydraulic manipulator without load. The reference signal is given by the black dashed line. At 5 seconds, the direction of the reference signal changes simultaneously, which is the situation with the highest power demand. The traditional PID cannot control the engine speed, so the performance under three different gear conditions is tested, while NMPC can adjust the engine speed. In addition, the flow efficiency is calculated according to \ref{eq14} and marked on the graph.}
	\label{noload}
\end{figure}

The proposed NMPC is first tested within a simulated environment, as it highlights the strengths and verifies safety. Fig.\ref{noload} presents the control performance under no-load conditions, compared with the PID control with dead zone compensation, which is most widely applied in hydraulic manipulators. 

Using a low gear setting maximizes energy efficiency, although this comes at the expense of a slower response time. In contrast, a high gear setting may enhance speed but lead to significant energy consumption. The NMPC method effectively combines control performance with energy efficiency, optimizing power usage only when the hydraulic manipulators are in motion. The NMPC method effectively combines control performance with energy efficiency, optimizing power usage only when the hydraulic manipulators are in motion. According to the principles of a positive flow system, the hydraulic pump stops outputting any flow, thereby preventing wastage. Additionally, we evaluate the performance of NMPC under two different conditions: one where the average value of the control sequence is used as input and another where the first value is used. The results indicate that using the average value leads to more stable control outcomes.

\begin{figure}[!t]
	\centering
	\includegraphics[width=3.4in]{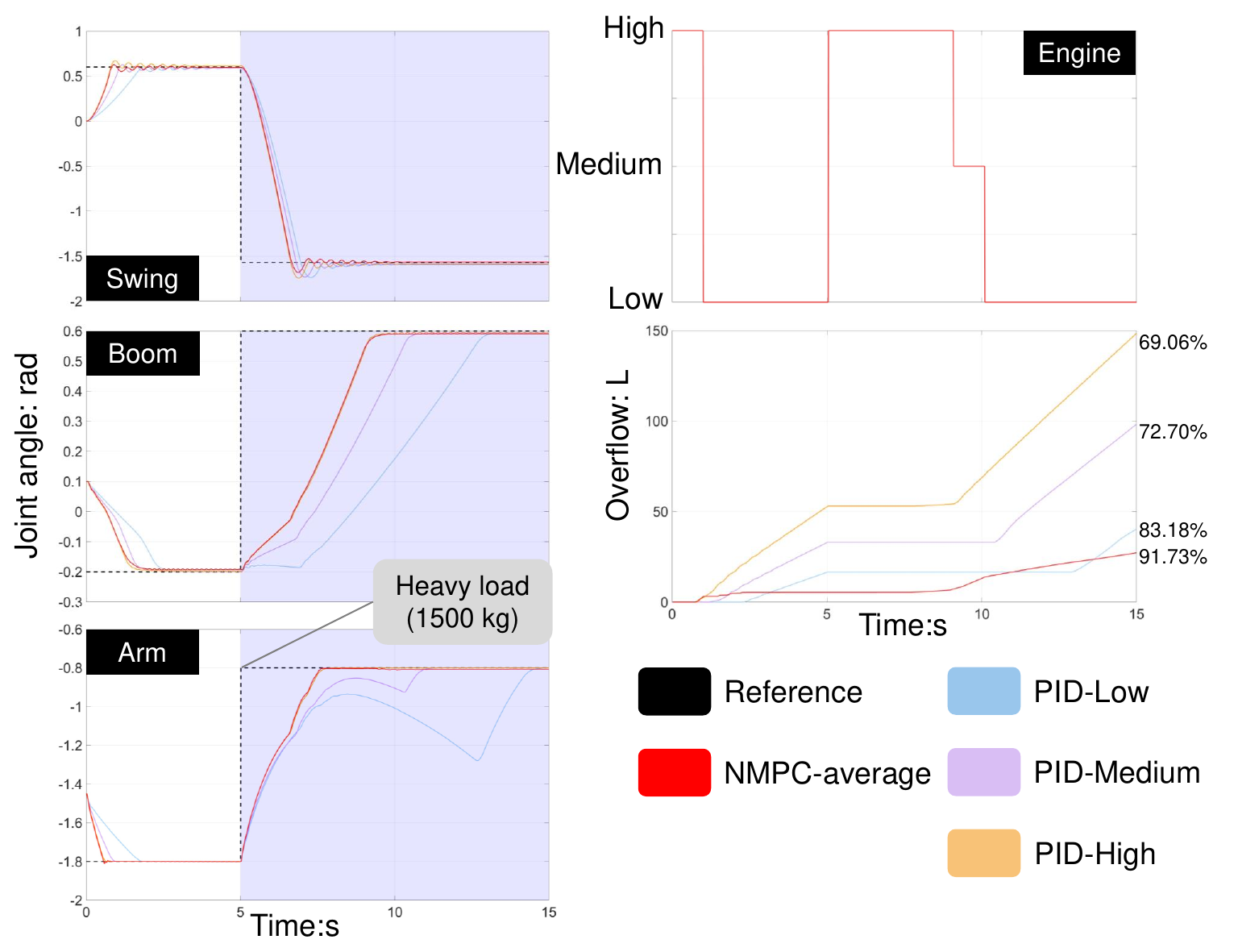}
	\caption{Comparison between the NMPC and a PID controller for the hydraulic manipulator under load. The direction of the signal  is changed concurrently with the introduction of a heavy load at 5s. Faced with inadequate engine power, the PID controller experiences a loss of control. In contrast, NMPC is capable of attaining performance similar to that of the PID controller under high-gear conditions and substantially improves the flow efficiency.}
	\label{load}
\end{figure}

We consider environmental interactions and introduce a load into the real-time control system. In Fig.\ref{load}, the results are presented. The arm joint offsets the desired reference trajectory due to the load when the gear is set to low and medium settings. Specifically, the system allocates output flow preferentially to the arm actuator, which has a minimal load, allowing it to move effectively under the load. However, when the boom joint begins to move, it takes a portion of the flow, leading to a deficit of hydraulic force in the arm's cylinder and causing a reverse flow. As the swing and boom approach their target positions, the flow supply is mostly redirected back to the arm actuator, enabling it to return to its original controlled state. The loss of control is especially hazardous in scenarios involving unmanned operations. While modifying the hydraulic circuit can reduce this risk, a practical approach in real-world applications requires proactive measures to prevent such events. Our method demonstrates robust performance even under load conditions. When environmental disturbances alter the dynamics model, the NMPC initially uses the offline predictive model. This approach helps prevent additional oscillations during the online model's learning process, enhancing safety. Once the online model compensates for any mismatches, the NMPC executes stable control.

\begin{figure}[!t]
	\centering
	\includegraphics[width=3.5in]{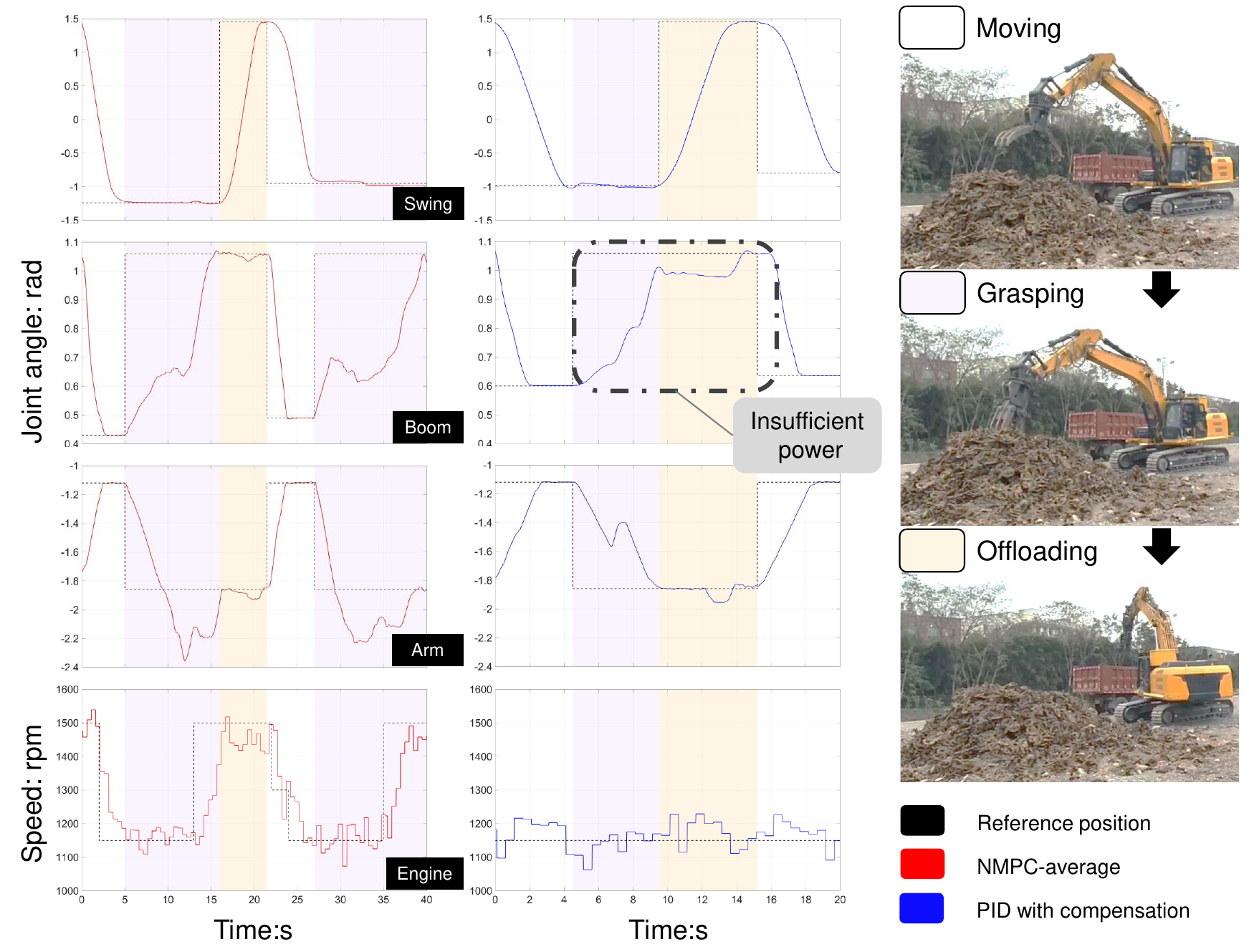}
	\caption{Experiments on a 22-ton hydraulic excavator. The tasks are mainly divided into moving, grasping, and offloading. The black dashed line represents the reference signal, which means that before grasping, the robot needs to reach the safe point, and after grasping, it needs to reach the truck point. The grasping process is controlled by visual servo. At low gear, the commonly used PID with compensation method performs poorly and cannot adapt to the dynamic model under load.}
	\label{grasp}
\end{figure}

After the simulation, the grasping task is conducted using a 22-ton hydraulic excavator, as shown in Fig.\ref{grasp}. This process can be divided into three stages: moving, grasping, and offloading. Firstly, the perception system detects the position of the scrap steel and provides the coordinates for the grasping point, prompting the excavator to move close to this location. In the second stage, the visual servo system engages to grasp the scrap steel and carefully adjusts it to a secure position. Finally, the coordinates of the placement point, transmitted by the perception system, guide the excavator to the target destination while under load, allowing it to release the grapple and offload the cargo. The visual servo module manages the grasping operations to ensure consistent quality of the scrap steel each time. Meanwhile, the remaining movements and unloading processes are controlled by NMPC. The compared method is the classic PID with compensation \cite{deng2017robust} at low gear settings. The inability to promptly adjust the output power resulted in power insufficiency-induced oscillations of the boom after an increase in load, and the boom can not achieve the target position until the load is removed. The experimental results of the NMPC method are consistent with the simulation, demonstrating the ability to regulate the speed of the diesel engine in real-time according to demand, providing an adequate flow during the motion process. Particularly in continuous engineering tasks where the load weight is not detectable, the NMPC method exhibits distinct advantages, offering a robust control strategy that can adapt to variations in load, thereby maintaining system stability and efficiency.

\subsection{Computational Complexity Analysis}\label{Results and Discussions-D}
Robots' real-time control requires a certain frequency of motion commands, thus minimizing the computation time for each NMPC iteration. Since the predictive model is SSMP, the computational complexity of the neural network is influenced by LSTM cycles, input dimensions, layers, neurons, activation function, and prediction length. Assuming a single layer LSTM cycle count of $h$, state dimension of $n$, and hidden neurons of $j$, the operations include the input gate, forget gate, and output gate, as well as the computations for the hidden states and cell units. The computational complexity can be approximated as $O(4jh(n+j+3))$ . The MLP has an input signal dimension of $m$, hidden neurons of $j$, with a single hidden layer, and a predictive horizon $N$. The process involves multiple matrix multiplications, and the computational complexity can be approximated as $O((m+1)j+(j+1)j+(N+1)j)$. Thus, the computational complexity of the entire framework can be approximated as follows:
\begin{equation}\label{eq15}
	\begin{aligned}
		O(4jh(n+j+3))+O((hn+2m+3j+2N+6)j)
	\end{aligned}
\end{equation}

\begin{table}[t]
	\begin{center}
		\caption{Computational Duration $[ms]$}\label{Computational Duration}
		\setlength{\tabcolsep}{1.6mm}
		\begin{tabular}{cccc}
			\toprule
			Parameters					& Prediction 	& Online model update 	& NMPC\\
			\midrule
			$h=20, j=128, N=10$ 		& 0.53			& 0.88					& 15.4 \\
			$h=20, j=64, N=10$        	& 0.34			& 0.41					& 10.4 \\
			$h=10, j=128, N=5$        	& 0.26			& 0.69					& 8.7 \\
			$h=10, j=64, N=5$        	& 0.15			& 0.22					& 5.5 \\
			\bottomrule
		\end{tabular}
	\end{center}
\end{table}

The number of LSTM cycles $h$ and the hidden neurons $j$ significantly impact the overall computational complexity. We implement the NMPC using an m-file in MATLAB 2022b with an Intel Core i9-13900HX CPU (laptop computer). We assess the computational time of the NMPC across different parameters, including the number of LSTM cycles $h$, hidden neurons $j$, and the predictive horizon $N$. This analysis includes the time taken for a single prediction, a single online model update (10 loops), and a single NMPC (30 iterations). Each test case is repeated 1000 times to ensure accuracy, and the results are summarized in Table \ref{Computational Duration}. The computational duration is essentially sufficient to meet the requirements of our engineering tasks (50Hz), and the frequency of the control signal can be increased by reducing hidden neurons, shortening the predictive horizon, or decreasing NMPC iterations.

\section{Conclusions}\label{Conclusions}
In response to the challenges of automation in unmanned industrial robotics, this paper proposes a data-driven NMPC framework that primarily addresses 1) accurate predictive models of nonlinear dynamics, 2) robot-environment interactions, and 3) universal solution methods. The proposed NMPC approach allows for customizing cost functions in convex form, aligning with various engineering tasks particularly adept at handling energy management requirements. This holds significant promise for industrial applications, as it stands to enhance economic performance.

First, we introduce the design process of the SSMP model. Based on Taylor expansion, the predictive horizon can be obtained from historical states and input sequences, prompting us to employ LSTM and MLP to construct an incremental predictive model. The proposed model outperforms standard MLP in multi-step prediction. Safety is considered by adopting a combination of open-loop and closed-loop methods in data collection, with the range determined by work tasks. Furthermore, to address model mismatches caused by the robot's interaction with the external environment, an online model is introduced to learn the variation. The results demonstrate that this online model significantly enhances the robot's predictive accuracy post-engagement with loads. This hybrid offline-online strategy also mitigates the risk of losing control. Finally, we choose a GD-based NMPC solution method and set an adaptive learning rate to mitigate the impact of model prediction errors. Such an optimization method applies to most convex function forms of cost functions and is highly flexible.

We conduct simulations and experiments on a 22-ton hydraulic excavator, employing a cost function to simultaneously control the multiple joints and the diesel engine. The results show that by adjusting the excavator's output power, we can increase fuel economy and safety. It is worth noting that hydraulic excavators encompass both power and actuation systems, and our approach can be extended to general industrial systems that involve energy generation and utilization. It is capable of addressing nonlinear challenges and achieving ideal energy distribution.

Currently, our approach is purely data-driven, without considering the impact of complex noise, and has yet to undergo detailed theoretical validation. The future works encompass the following: 1) developing a hybrid predictive model that combines data with physics to reduce the burden of data collection, 2) addressing data processing in the presence of complex noise, 3) validating simulation and experimental results through comprehensive theoretical analysis, and 4) transforming our approach into commercial software.





 
%

\bibliographystyle{IEEEtran}
\bibliography{IEEEfull,Ref2}

\vfill

\end{document}